\documentclass{article}

\usepackage[preprint]{corl_2026} 
\usepackage{ulem}
\usepackage{booktabs}
\usepackage{bm}
\usepackage{caption}
\usepackage{graphicx}
\usepackage{enumitem}
\usepackage{wrapfig}
\usepackage{graphicx}
\usepackage{multirow}
\usepackage{float}
\usepackage{wrapfig}
\usepackage{amssymb}
\usepackage{threeparttable}
\usepackage{listings}
\usepackage{xcolor}

\title{A Practical Recipe Towards Improving Sim-and-Real Correlation for VLA Evaluation 
}

\author{
  Shuo Wang$^{1,2*}$ \quad Hanyuan Xu$^{1*}$ \quad Yingdong Hu$^{1,2}$ \quad Fanqi Lin$^{1,2}$ \quad Yang Gao$^{1,2\dag}$\\
  $^1$Tsinghua University \quad $^2$Shanghai Qi Zhi Institute\\
  \texttt{runner21st@gmail.com} \quad \texttt{gaoyangiiis@mail.tsinghua.edu.cn} \\
  $^*$Equal contribution \quad
  $^\dag$Corresponding Author
}

\begin{document}
\maketitle


\begin{abstract}
    Simulation has become an essential tool for evaluating and improving vision-language-action (VLA) policies, offering scalable, reproducible, and controllable alternatives to costly real-world robot evaluation. Recent simulation benchmarks have made substantial progress on realism and diversity, yet these platforms have not been widely adopted as reliable proxies for real-world policy evaluation. In this work, we investigate this issue through the lens of sim-and-real correlation. We conduct a systematic study across multiple simulation platforms, VLA policies, tasks, and perturbation factors, measuring whether simulated evaluation preserves real-world conclusions in terms of policy ranking consistency, performance correlation, and perturbation-wise failure patterns. This analysis allows us to characterize the limitations of existing simulators and identify what kinds of simulation signals are more aligned with real-world deployment. We further examine how users should exploit simulation for policy improvement, including when simulator-based finetuning is beneficial and how the amount of post-training data affects sim-and-real alignment. Overall, our work provides a unified framework for measuring, interpreting, and improving the usefulness of simulation for VLA policies, offering guidance both for simulator designers and for practitioners who use simulation as part of the policy development pipeline.
\end{abstract}



\section{Introduction}

Embodied artificial intelligence has recently seen rapid progress toward generalist robot policies that can act across diverse tasks, scenes, objects, and embodiments~\cite{brohan2022rt1,brohan2023rt2,ghosh2024octo,kim2024openvla,black2024pi0,sapkota2025vision}. As the number of such models grows, evaluation has become a central bottleneck~\cite{openx2023,khazatsky2024droid,zhou2025autoeval}. Reliable real-world evaluation requires many rollouts across different task configurations, scene appearances, and object variations, making it expensive, slow, and difficult to scale \cite{tri2025lbm,atreya2025roboarena}. 

To alleviate this bottleneck, recent work has devoted substantial effort to building simulation benchmarks with greater task diversity, visual and physical realism, and structured perturbations \cite{zhu2020robosuite,nasiriany2024robocasa,zhang2024vlabench}. These benchmarks provide scalable platforms for probing the capabilities and limitations of robot policies \cite{gu2023maniskill2,james2020rlbench}. More recent benchmarks further adopt real-to-sim techniques, aiming to reduce the gap between simulator and real world, thereby improving the correlation between simulated and real-world evaluation results~\cite{li2024simpler, kadian2020sim2real}.

Yet a more basic question remains unresolved: \textbf{when does simulation actually act as a \textit{\emph{measurement proxy}} for real-world policy behavior?} A simulator may contain many tasks, look visually realistic, or produce challenging tasks, but these properties do not guarantee the policy-level conclusions drawn from simulator evaluation could be aligned with the physical world \cite{chebotar2018simopt}. For a policy developer, the practical value of using simulation is measured by whether it preserves the decisions that matter: which policy performs better, which perturbation dimensions expose greater model sensitivity, and so on \cite{atreya2025roboarena,li2024simpler,sedlacek2025realm}. This leads to two questions: For simulator designers, what properties should a simulator have to preserve real-world policy-level conclusions? For simulator users, how should a given simulator be used or adapted to further improve its sim-and-real correlation?

In this work, we conduct a systematic study for simulators through the lens of sim-and-real correlation. Rather than proposing another new simulator, we fairly evaluate five VLA policies across different simulation platforms, enabling direct comparison between model rankings in simulation and on real world evaluation \cite{li2024simpler,sedlacek2025realm,zhang2025vlaarena}. Specifically, we compare simulated and real-world evaluation under vision, object-layout, language, and behavior perturbations (i.e. changes to unseen objects). Our results verify the importance of simulator-level visual fidelity, and further show that reliable simulators should make policies fail in ways similar to the real world. For example, if a policy is more vulnerable to object-layout changes on real robots, the simulator should reveal the same weakness, rather than making other perturbations appear relatively easier or more difficult.

We further study how users should exploit simulators to improve sim-and-real correlation: should pretrained policies be directly evaluated by the simulator in a zero-shot manner, or should they be first fine-tuned before evaluation~\cite{kim2025finetuningvla,tian2025interndataa1}? Furthermore, by varying the amount of data used for fine-tuning, we analyze how tuning data amount affects the sim-and-real correlation. Results confirm that fine-tuning within an adequate range can make model behavior in simulation more closely resemble that in the real world. Overall, our contributions are three-fold:

\begin{itemize}
    \item  We conduct a comprehensive sim-and-real evaluation across multiple simulation platforms, VLA policies, manipulation tasks, and perturbation dimensions. Through this unified evaluation, we systematically examine the correlation between existing simulators and real-world evaluation, and further study whether these simulators can reveal model sensitivity to perturbations along different dimensions.

    \item We derive practical insights for selecting and \textbf{designing} reliable simulators for VLA evaluation. Our analysis shows that simulator proxies should be visually and physically realistic, and should reflect how policies fail under different perturbations in the real world. This suggests that perturbations should be designed to match the relative difficulty of real-world evaluation across vision, layout, language, and behavior variations.

    \item  We further provide user-oriented guidance for \textbf{using} simulation in VLA policy evaluation. In particular, we show that fine-tuning can improve the sim-and-real correlation, but this benefit is not monotonic with data scale: post-training with an adequate amount of data could strengthen proxy reliability, whereas excessive fine-tuning may cause the model fit to fixed behavior trajectories, thereby obscuring its true generalization ability under distribution shifts.
    
\end{itemize}


\section{Related Work}

\subsection{Simulation Benchmarks for Embodied and VLA Evaluation}

Simulation has become a standard tool for scalable and reproducible evaluation of embodied agents and robot policies~\citep{james2020rlbench, mees2022calvin, liu2023libero}. Early benchmarks such as ALFRED, Habitat 2.0, and BEHAVIOR-1K introduced interactive household environments with language instructions, navigation, object interaction, and long-horizon tasks~\citep{shridhar2020alfred, szot2021habitat, li2024behavior1k}. Manipulation benchmarks including RLBench, CALVIN, LIBERO, ManiSkill, and RoboSuite further standardized language-conditioned manipulation, skill composition, lifelong learning, and policy evaluation in controlled physical environments~\citep{zhu2020robosuite, gu2023maniskill2}.

Recent benchmarks have improved simulation along three directions: realism, scale, and support for evaluating generalist robot policies. RFUniverse and ARNOLD focus on richer physics, multiphysics interaction, and continuous object-state grounding~\citep{fu2022rfuniverse, gong2023arnold}; RoboCasa scales everyday manipulation with diverse scenes, objects, embodiments, and synthetic trajectories~\citep{nasiriany2024robocasa, nasiriany2026robocasa365}; while VLABench and VLA-Arena introduce structured difficulty and controlled language or visual perturbations for VLA evaluation~\citep{zhang2024vlabench, zhang2025vlaarena}. In parallel, some recent efforts like EmbodiedBench and ManipBench evaluate MLLMs as embodied planners or reasoning modules~\citep{choi2024lota, liu2024visualagentbench, cheng2025embodiedeval, yang2025embodiedbench, luo2025robobench, zhao2025manipbench}. While these works mainly measure capability within benchmarks, our work asks whether simulation preserves real-world conclusions for model selection and post-training design.

\subsection{Sim-and-Real Correlation and Simulator-Based Policy Improvement}

Another line of work studies whether simulation can approximate real-world robot performance. SIMPLER identifies visual and control gaps as key obstacles and builds simulation settings correlated with real-world manipulation behavior~\citep{li2024simpler}. REALM validates VLA generalization with high-fidelity visuals, aligned control, multiple manipulation skills, and controlled perturbations~\citep{sedlacek2025realm}. Large-scale real-world studies further emphasize controlled distribution shifts, large trial counts, rigorous protocols, and statistically grounded relative comparisons~\citep{tri2025lbm, agarwal2021deep}. These works suggest that simulation can provide useful evaluation signals when visual distributions are well aligned.

Our work differs by comparing multiple simulation platforms through their ability to preserve real-world decisions, rather than validating a single simulator~\citep{li2024simpler, sedlacek2025realm}. We measure score correlation, policy-ranking consistency and perturbation sensitivity alignment. We also extend to post-training in simulators~\cite{xing2021kitchenshift}: Prior works such as RoboCasa and InternData-A1 show that synthetic or simulator-generated data can improve imitation learning and VLA pre-training~\citep{nasiriany2024robocasa,tian2025interndata}. Rather than treating simulator data only as a way to improve task performance, we study whether simulator-based adaptation can make simulated evaluation better aligned with real-world behavior, as measured by policy-ranking consistency and perturbation-sensitivity alignment.

	

\section{Evaluation Platform}

\begin{figure}[h]
    \centering
    \includegraphics[width=\linewidth]{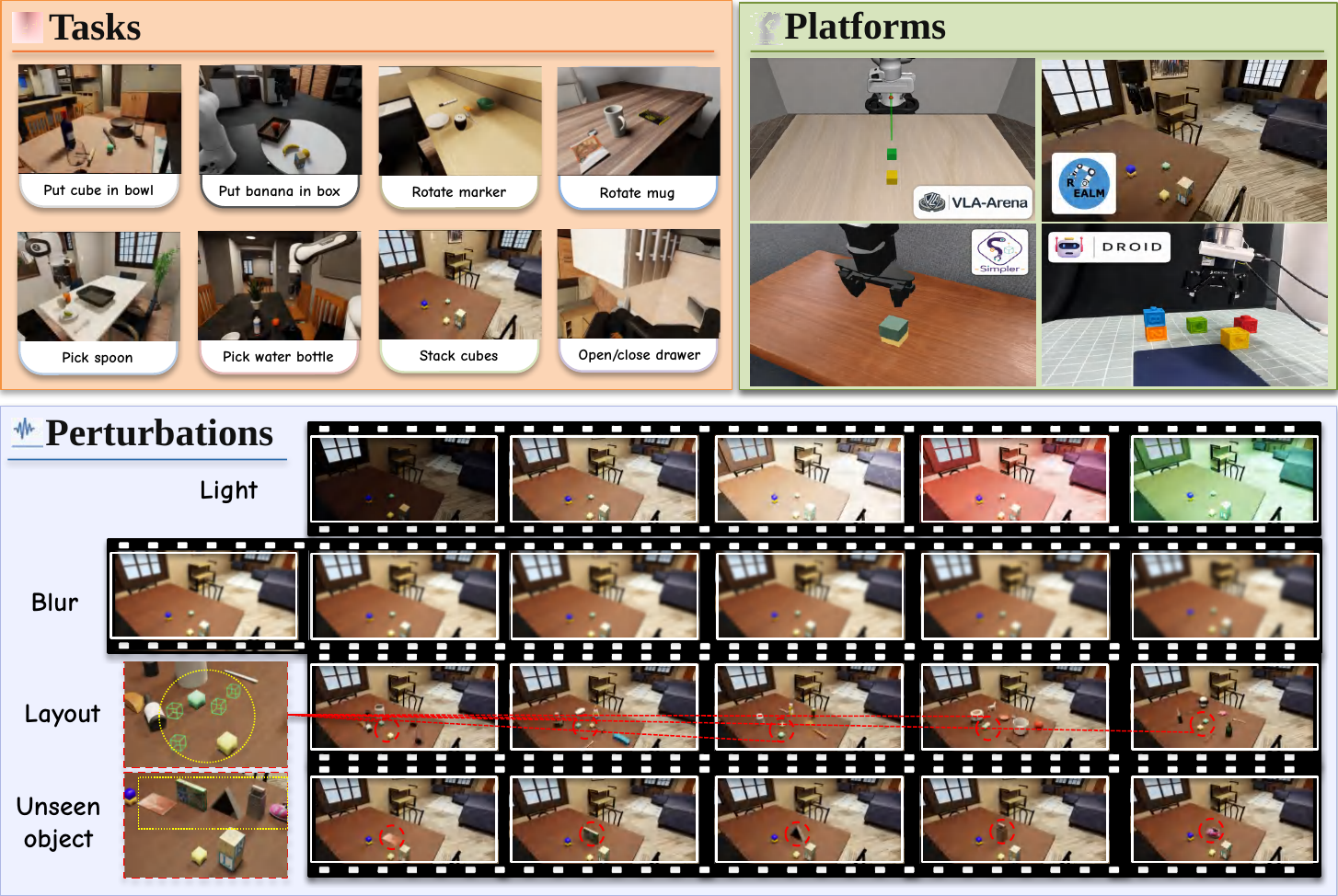}
    \caption{Overview of our unified sim-and-real evaluation platform. We align tabletop manipulation tasks across simulation and real-world settings, and evaluate them under controlled perturbations.}
    \label{platform}
\end{figure}

To systematically measure the simulations for VLA policy evaluation, we construct a unified evaluation platform spanning nine tabletop manipulation tasks as shown in Figure \ref{platform}. These tasks are designed to cover a broad range of action spaces and interaction patterns, including pick-and-place, object rotation, opening and closing articulated objects, and fine-grained manipulation. For each task, we introduce controlled perturbations along four complementary dimensions: \textit{vision}, \textit{language}, \textit{layout}, and \textit{behavior}. Vision perturbations modify the visual appearance of the scene, including blur and contrast, light color and conditions; language perturbations change the form of task instructions, layout perturbations alter spatial configurations, and behavior perturbations evaluate generalization to unseen manipulated objects. Detailed visualizations of tasks and perturbations are shown in Appendix~\ref{taskvis} and \ref{pertvis}.

We further build or select tasks whose action spaces, visual appearances and success objectives are closely aligned with their real-world counterparts for fairness. Rather than directly comparing arbitrary benchmark tasks, we emphasize task-level correspondence: each simulated task is matched to the same underlying manipulation skill and completion criterion, so that differences in evaluation results are more likely to reflect simulator fidelity rather than mismatched task definitions. We instantiate the same nine-task suite on real hardware using the DROID embodiment~\cite{khazatsky2024droid}, enabling direct comparison between simulated and real-world policy behavior. This paired sim-and-real design allows us to study whether simulation preserves the conclusions that matter for policy development.

\section{Experimental Results}
\label{sec:result}

We organize our experiments around two research questions. First, we ask whether existing simulated VLA benchmarks can faithfully predict real-world model rankings and robustness patterns. This analysis allows us to identify which type of simulator can provide more reliable real-world proxy signal. Second, based on the selected simulator, we study whether targeted simulator-based post-training can further improve real-world performance and sim-real evaluation alignment.

\subsection{Implementation details}

All results reported in this section are evaluated with $20$ simulated rollouts and $5$ real-world rollouts for each policy, task, and perturbation dimension. In total, our evaluation includes 11,800 simulated rollouts and 1,115 real-world rollouts. To ensure fair comparison across benchmarks, for newly constructed tasks in simulation, we collect task-specific data through teleoperation and fine-tune all evaluated policies using the same collected data. This controlled protocol ensures that differences in evaluation results are not caused by unequal access to adaptation data, making the measurements across benchmarks directly comparable.

\subsection{Are Existing VLA Benchmarks Faithful Predictors of Real-World Model Ranking?}
\label{proxy}

We first compare three representative simulated VLA benchmarks: VLA-Arena, SIMPLER, and REALM across five recent VLA policies: $\pi_0$, $\pi_0$-FAST, $\pi_{0.5}$, GR00T N1.6, and GR00T N1.7. For compact notation in the tables, we denote these models as $A,B,C,D,E$ respectively.

\begin{table}[htbp]
\centering
\caption{
Policy rankings across simulated benchmarks and real-world evaluation. Policies are ordered in descending performance.
}
\label{tab:policy_rankings}
\setlength{\tabcolsep}{5pt}
\begin{tabular}{cccccc}
\toprule
\textbf{Benchmark} & \textbf{Simulator} & \textbf{Vision} & \textbf{Layout} & \textbf{Language} & \textbf{Behavior} \\
\midrule
VLA-Arena  & Mujoco & CBAED & CEADB & CABED & CABED \\
SIMPLER    & Sapien & EDCAB & ECDAB & CEDAB & -- \\
REALM   & Isaac-Sim    & CEBAD & CEABD & CEBAD & CEBAD \\
\midrule
Real World & -  & ECADB & CAEBD & CABED & CEADB \\
\bottomrule
\end{tabular}
\end{table}


\paragraph{Results.}
Table~\ref{tab:policy_rankings} shows the policy rankings under each benchmark and perturbation dimension.
The rankings are sorted from best to worst.
A faithful simulator should preserve these rankings, because model selection is fundamentally a relative decision:
if one policy outperforms another in the real world, a reliable simulator should ideally preserve the same ordering. To quantify ranking consistency, Table~\ref{tab:real_to_sim_metrics} reports Spearman rank correlation ($\rho$), Pearson correlation ($r$), and Mean Maximum Rank Violation (MMRV) between each simulator and real-world evaluation. Spearman correlation measures whether the simulator preserves the relative model ordering. Furthermore, Pearson correlation measures whether simulator scores linearly track real-world scores, while MMRV measures the severity of high-margin ranking mistakes~\cite{li2024simpler}; lower MMRV means the simulator is less likely to severely mislead model selection. Detailed computation formulas for these metrics are provided in Appendix~\ref{metrics}.

\paragraph{Observation 1: Reliable simulators should make policies fail like the real world.}
We first study the simulator-design question: what properties make a simulator a reliable proxy for real-world VLA evaluation? For a simulator to provide actionable evaluation signals, its perturbations should also stress policies in a realistic way. In other words, perturbations that cause large performance degradation in the real world should also be severe in simulation, while mild real-world perturbations should not dominate simulated evaluation. Therefore, beyond policy-ranking proxy ability, we further measure whether simulated perturbations induce realistic vulnerability patterns.

For each environment $e$, policy $m$, and perturbation dimension $d$, let $S_{e,m,d}$ denote the success rate.
We define the normalized perturbation sensitivity as
\begin{equation}
I_{e,m,d}
=
\frac{
\max_{d' \in \mathcal{D}_e} S_{e,m,d'} - S_{e,m,d}
}{
\max_{d' \in \mathcal{D}_e} S_{e,m,d'} - \min_{d' \in \mathcal{D}_e} S_{e,m,d'}
},
\end{equation}
where $\mathcal{D}_e$ is the set of perturbation dimensions supported by environment $e$.
This normalization removes the absolute success-rate scale and focuses on each policy's relative vulnerability pattern.
A value of $0$ means that the perturbation is the least damaging dimension for that policy, while a value of $1$ means that it is the most damaging one.

\begin{wraptable}{r}{0.53\linewidth}
\centering
\caption{
Sim-and-real correlation between each simulated benchmark and real-world evaluation.
}
\label{tab:real_to_sim_metrics}
\vspace{0.5em}
\resizebox{\linewidth}{!}{
\begin{tabular}{ccccc}
\toprule
\textbf{Env.} & \textbf{Dim.} & $\boldsymbol{\rho}\uparrow$ & $\boldsymbol{r}\uparrow$ & \textbf{MMRV}$\downarrow$ \\
\midrule
\multirow{5}{*}{VLA-Arena}
             & Vision   & 0.000 & 0.241 & 0.188 \\
          & Layout   & 0.800 & 0.858 & 0.016 \\
          & Language & 1.000 & 0.987 & 0.000 \\
          & Behavior & 0.500 & 0.814 & 0.035 \\
          & Avg.     & 0.575 & 0.725 & 0.060 \\
\midrule
\multirow{4}{*}{SIMPLER}
            & Vision   & 0.700 & 0.701 & 0.102 \\
          & Layout   & 0.300 & 0.420 & 0.056 \\
          & Language & 0.200 & 0.086 & 0.226 \\
          & Avg.     & 0.400 & 0.402 & 0.128 \\
\midrule
\multirow{5}{*}{REALM}
            & Vision   & 0.600 & 0.672 & 0.046 \\
          & Layout   & 0.900 & 0.865 & 0.008 \\
          & Language & 0.600 & 0.722 & 0.042 \\
          & Behavior & 0.700 & 0.882 & 0.023 \\
          & Avg.     & \textbf{0.700} & \textbf{0.785} & \textbf{0.030} \\
\bottomrule
\end{tabular}
}
\end{wraptable}

\begin{figure}[t]
    \centering

    \begin{minipage}[t]{0.49\linewidth}
        \vspace{0pt}
        \centering
        \captionof{figure}{Relation between policy-ranking proxy ability and perturbation-sensitivity alignment.}
        \label{image1}
        \vspace{2pt}
        \includegraphics[width=\linewidth]{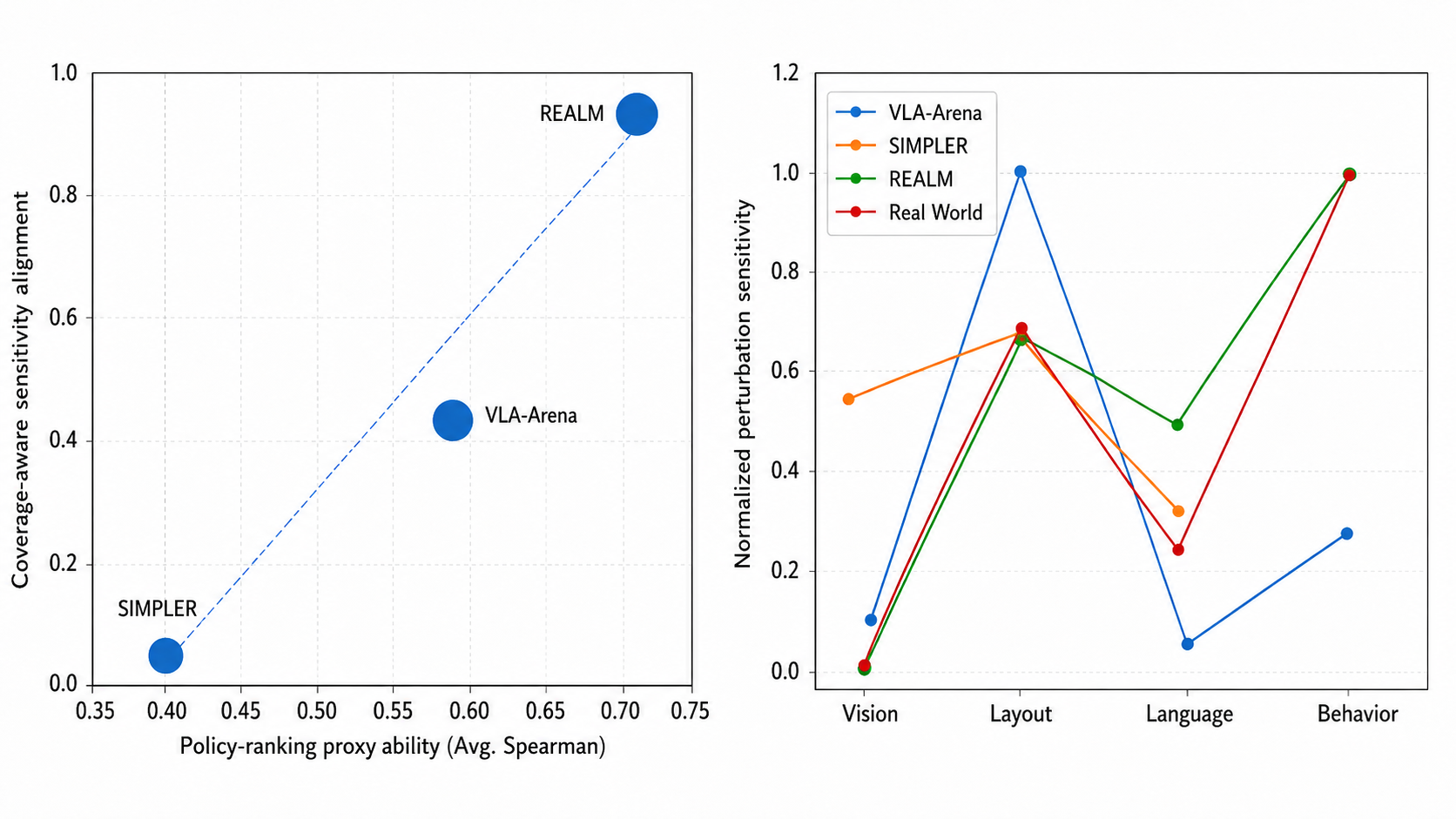}
    \end{minipage}
    \hfill
    \begin{minipage}[t]{0.49\linewidth}
        \vspace{0pt}
        \centering
        \captionof{table}{
        Average normalized perturbation sensitivity profiles across policies.
        }
        \label{tab:perturbation_sensitivity_profile}
        \vspace{2pt}

        \setlength{\tabcolsep}{3pt}
        \renewcommand{\arraystretch}{0.95}

        \resizebox{\linewidth}{!}{%
        \begin{tabular}{ccccc}
        \toprule
        \textbf{Env.} & \textbf{Vision} & \textbf{Layout} & \textbf{Language} & \textbf{Behavior} \\
        \midrule
        VLA-Arena & 0.095 & 1.000 & 0.031 & 0.297 \\
        SIMPLER   & 0.540 & 0.670 & 0.340 & -- \\
        REALM     & 0.000 & 0.644 & 0.473 & 1.000 \\
        \midrule
        Real World & 0.008 & 0.679 & 0.241 & 1.000 \\
        \bottomrule
        \end{tabular}%
        }
    \end{minipage}

\end{figure}

Figure~\ref{image1} reveals a clear positive association between policy-ranking proxy ability and perturbation-sensitivity alignment. Simulators that better preserve real-world policy rankings also tend to produce perturbation-induced failures that are more consistent with the real world. This relationship is important for simulator design: a simulator becomes a better proxy not merely by including more tasks or more perturbation types, but by making those perturbations affect policies with realistic severity. Thus, perturbation realism should be understood not only as visual or physical plausibility, but also as whether the induced policy failures match real-world deployment.



Table~\ref{tab:perturbation_sensitivity_profile} further shows the average normalized sensitivity profile across policies.
In the real world, behavior perturbations are the most damaging, followed by layout perturbations, while vision and language perturbations are relatively less severe.
REALM is the only evaluated simulator that preserves this perturbation hierarchy across all four dimensions:
vision is the least damaging, behavior is the most damaging, and layout lies between them.
By contrast, VLA-Arena over-emphasizes layout perturbations relative to behavior perturbations, while SIMPLER omits behavior perturbations entirely.



These results suggest a concrete design principle:
\emph{a trustworthy simulator should not only rank policies like the real world, but also make them fail like the real world.} Simulator perturbations should therefore be designed to induce vulnerability patterns consistent with real-world deployment. Benchmarks that reliably reflect real-world performance are valuable not merely as tools for ranking models, but also as diagnostic instruments for understanding failure modes and guiding future model improvements.


Based on these findings, we use REALM as the primary simulator for our post-training study in Section~\ref{sec:post_training}.
Our goal in the next section is not merely to improve simulated performance, but to investigate whether targeted post-training can improve policy performance and further strengthen sim-real evaluation alignment.

\paragraph{Observation 2: Proxy gaps are dominated by simulator-level fidelity rather than object-level task mismatch.}

A remaining concern is that the differences observed across simulators may be caused by imperfect task alignment rather than by simulator fidelity itself. Even when two simulators implement the same high-level task, the manipulated object, object geometry, grasp affordance, or success condition may differ slightly.

\begin{wrapfigure}{r}{0.4\linewidth}
    \centering
    \includegraphics[width=\linewidth]{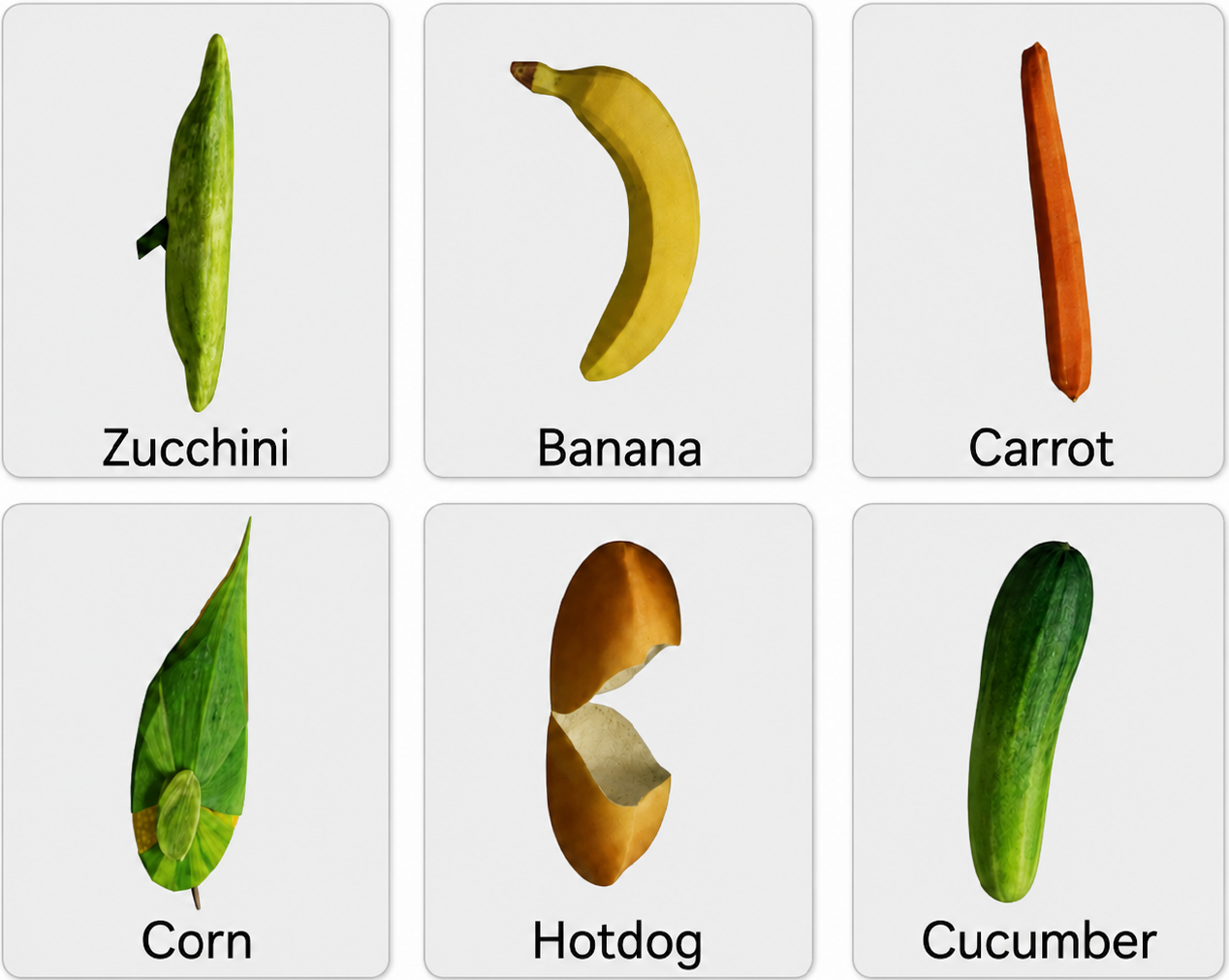}
    \caption{Visualization of different objects in REALM.}
    \label{Objects}
\end{wrapfigure}

To address this concern, we conduct an ablation study in REALM on the task \textit{pick the banana into the box}, which exhibits the lowest cross-simulator alignment among our selected tasks. This is mainly because the manipulated objects to be grasped differ across simulators. We replace the banana with five new manipulated objects in Figure \ref{Objects}: corn, zucchini, hotdog, carrot, and cucumber. For each new object, we collect a small amount of object-specific data and perform fine-tuning with few steps.

We find that object replacement changes the absolute success rates of different policies under the default setting. However, the relative policy ordering remains \textbf{unchanged} across all object variants. In other words, changing the manipulated object within the same simulator affects how well policies perform, but is insufficient to induce rank inversions among them. This leads to our second observation:
\emph{the proxy ability of a simulator is primarily governed by simulator fidelity rather than residual object-level task mismatch.} For simulator design, this implies that simulators should be realistic, as simulator-intrinsic properties largely determine sim-and-real correlation.

\subsection{Does Targeted Post-Training Improve Sim-Real Evaluation Alignment?}
\label{sec:post_training}

The previous section studies the simulator-design question: what properties make a simulator a reliable proxy for real-world evaluation? In this section, we turn to the user-facing question: once a simulator is available, how should it be used to obtain reliable evaluation signals? 

This question is important because VLA policies are typically pretrained on broad and heterogeneous data distributions, while any particular simulator defines a much narrower environment distribution through its rendering style, object assets, camera configuration, contact dynamics, and so on. As a result, directly evaluating a pretrained policy in simulation may conflate two factors: the policy's true real-world capability and its adaptation gap to the simulator. A simulator may therefore be a better proxy after the policy has been lightly adapted to its environment, rather than under a purely zero-shot evaluation protocol.

\begin{table}[h]
\centering
\caption{
Policy-ranking proxy ability and perturbation-sensitivity alignment before and after fine-tuning.
}
\label{tab:proxy_sensitivity_finetune}
\setlength{\tabcolsep}{5pt}
\resizebox{\linewidth}{!}{
\begin{tabular}{cccccccc}
\toprule
\textbf{Env.} 
& \textbf{Dims.} 
& \textbf{Proxy} $\boldsymbol{\rho}\uparrow$ 
& \textbf{Proxy} $\boldsymbol{r}\uparrow$
& \textbf{Proxy MMRV}$\downarrow$
& \textbf{Sens.} $\boldsymbol{\rho}\uparrow$
& \textbf{Sens.} $\boldsymbol{r}\uparrow$
& \textbf{Sens. MAE}$\downarrow$
 \\
\midrule
Original & 4 & 0.700 & 0.785 & 0.030 & 0.924 & 0.875 & 0.110 \\
Finetuned  & 4 & \textbf{0.875} & \textbf{0.878} & \textbf{0.015} & \textbf{0.955} & \textbf{0.970} & \textbf{0.041} \\
\bottomrule
\end{tabular}
}
\end{table}

\paragraph{Observation 3: Simulator-based post-training significantly improves sim-and-real correlation.}
We conduct targeted post-training in REALM, which provides the strongest proxy signal among the evaluated simulators in Section~\ref{proxy}.
For each task, we collect a small amount of simulator data from the target evaluation distribution and fine-tune the policy for a limited number of steps.
We then re-evaluate the fine-tuned policies under the same perturbation dimensions and compare the resulting simulated measurements with real-world results.

Table~\ref{tab:proxy_sensitivity_finetune} shows that post-training consistently strengthens sim-and-real alignment.
Compared with the original zero-shot evaluation, fine-tuning improves the average policy-ranking Spearman correlation from $0.700$ to $0.875$, while reducing Proxy MMRV from $0.030$ to $0.015$.
This indicates that post-training makes the simulator less likely to produce high-margin ranking mistakes when used for model selection.
The improvement is not limited to policy ranking: perturbation-sensitivity alignment also increases, with sensitivity Spearman correlation improving from $0.924$ to $0.955$ and Pearson correlation from $0.875$ to $0.970$.
Meanwhile, Sensitivity MAE decreases substantially from $0.110$ to $0.041$, suggesting that the fine-tuned simulator evaluation more accurately reflects the real-world severity of different perturbations.

\begin{wrapfigure}{r}{0.45\linewidth}
    \centering
    \includegraphics[width=\linewidth]{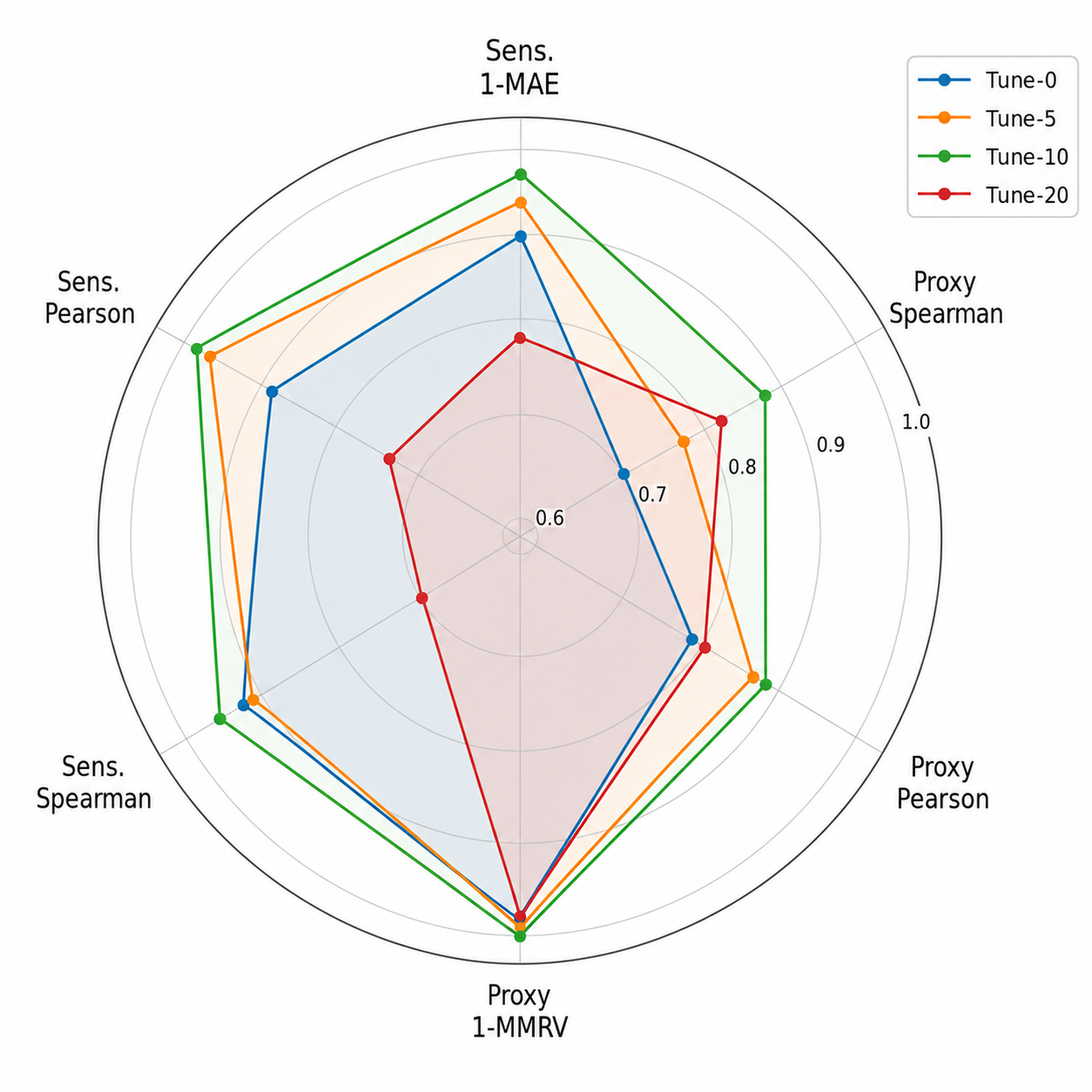}
    \caption{Results under different post-training data amounts.}
    \label{fig:finetune_amount_radar}
\end{wrapfigure}

It is worth noting that simulator-based post-training does not necessarily improve the absolute performance of every policy on every task. For example, on some tasks such as \textit{stack the cube}, $\pi$-series policies even exhibit lower success rates after fine-tuning than before. This observation highlights an important distinction: the purpose of collecting simulator data and performing fine-tuning is not simply to boost task performance. Rather, post-training should be viewed as an adaptation and calibration step that reduces the distribution gap between pretrained data and the simulator environment. Its main value lies in making simulated evaluation more predictive of real-world behavior, as reflected by stronger policy-ranking consistency and better perturbation-sensitivity alignment.

\paragraph{Observation 4: More simulator fine-tuning data does not monotonically improve proxy reliability.}

We further study how the amount of simulator-based post-training data affects sim-and-real correlation. While Observation~3 shows that lightweight post-training can substantially improve the reliability of simulation as an evaluation proxy, it remains unclear whether using more simulator data and longer fine-tuning always leads to better alignment.

To answer this question, we vary the amount of simulator-based fine-tuning and evaluate the resulting policies using the same proxy-ranking and perturbation-sensitivity metrics.  Figure~\ref{fig:finetune_amount_radar} shows that the relationship is non-monotonic, where Tune-$n$ means finetune with $n$ demos each task. Compared with original REALM, Tune-5 already improves several proxy metrics, indicating that even a small amount of simulator adaptation can reduce the distribution mismatch between pretrained policies and the simulator environment. Tune-10 achieves the best overall alignment, with the highest policy-ranking correlation and strongest perturbation-sensitivity alignment. However, further increasing the fine-tuning amount to 20 does not continue this trend. Although Tune-20 still achieves reasonable policy-ranking consistency, its perturbation-sensitivity alignment drops substantially, becoming even worse than non-tuned REALM. We attribute this mainly to the limited complexity of the simulated environment. When the amount of fine-tuning data becomes large, the model may overfit to the single-task distribution, making it less sensitive to perturbations such as language variations. As a result, its generalization performance on certain perturbation dimensions can degrade rather than improve.

This suggests that simulator-based post-training should be viewed as a calibration mechanism rather than a simple performance-scaling recipe. Light adaptation may leave a large sim-environment gap, while excessive adaptation can overfit policies to simulator-specific artifacts and distort how perturbations affect the policy. Therefore, for simulator users, the most reliable proxy signal is obtained not by maximizing simulator fine-tuning, but by choosing an intermediate adaptation regime that improves sim-and-real correlation without overfitting to the simulator.

\section{Conclusion, Limitation, and Future work}
\label{sec:conclusion}

In this work, we revisit the role of simulation in VLA policy evaluation from a measurement-oriented perspective. Through the lens of sim-and-real correlation, we study when simulators can serve as a reliable proxy for real-world policy behavior. Through a unified sim-and-real evaluation, our study shows that improving sim-and-real correlation requires both reliable simulator design and appropriate simulator use. Simulators should preserve real-world policy rankings and reflect how policies fail under different perturbations, while adequate fine-tuning can further strengthen the proxy reliability. Overall, our work provides a practical recipe towards improving sim-and-real correlation for VLA evaluation, and takes a step forward in building more reliable and useful simulation-based evaluation for generalist robot policies.

Although we conduct a broad set of experiments across multiple simulators, policies, tasks, and perturbation dimensions, our empirical analysis is still conducted within a bounded evaluation scope. Specifically, our real-world experiments focus on tabletop manipulation with a fixed embodiment, covering a finite set of policies and task suites. Therefore, the results should not be interpreted as a definitive ranking of existing simulators or policies. Such a conclusion would require substantially broader evaluation across more embodiments, task domains, interaction types, and real-world deployment conditions.

However, this limitation is aligned with the central goal of our work. 
Rather than claiming which simulator or policy is universally superior, we aim to study when simulation provides decision-relevant signals and how its usefulness can be measured through sim-and-real correlation. 
Our observations are intended to provide practical guidance for simulator designers and users. We believe that larger-scale and more comprehensive evaluations will further strengthen this line of analysis, leading to more stable and valuable conclusions.

\clearpage


\bibliography{example}  

\newpage
\appendix

\section{Visualization of aligned tasks}\label{taskvis}

\begin{figure}[h]
    \centering
    \includegraphics[width=\linewidth]{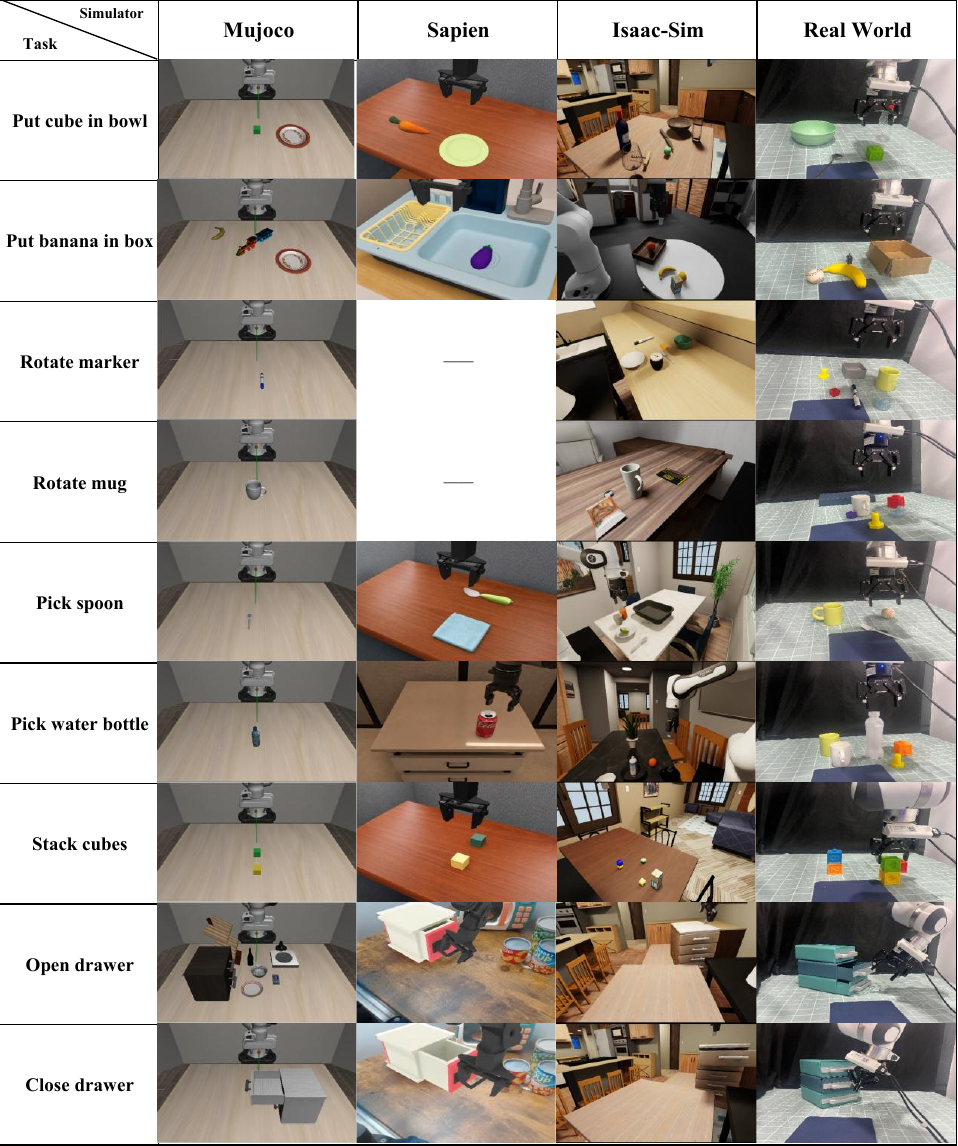}
    \caption{Visualization of aligned tasks.}
    \label{visualization}
\end{figure}

We visualize the nine aligned tasks in Figure~\ref{visualization}. Due to the limited availability of corresponding training data, two rotation tasks are not evaluated in SIMPLER.

\section{Detailed Related Work}

Simulation has been widely used to provide scalable and reproducible testbeds for embodied agents and robot policies~\cite{ahmed2020causalworld,ayub2021f,manivasagam2020lidarsim}. Beyond constructing realistic or diverse simulation benchmarks, a growing line of work studies whether simulation can serve as a reliable measurement proxy for real-world policy behavior~\cite{mittal2023orbit,maddukuri2025sim,torne2024reconciling, abou2025real}. Early embodied-AI studies on sim-to-real predictivity showed that simulated performance does not automatically imply real-world performance, and that simulator usefulness should be validated through paired sim-and-real measurements rather than assumed from visual realism alone~\citep{kadian2020sim2real}. In robotic manipulation, SIMPLER formalizes this perspective for policies trained on real data and evaluated in simulation, emphasizing that an effective simulator should preserve relative policy performance and robustness patterns under controlled visual and control gaps~\citep{li2024simpler}. This is complementary to recent real-world evaluation infrastructures such as AutoEval and RoboArena, which aim to make physical policy comparison more scalable and reproducible~\citep{zhou2025autoeval,atreya2025roboarena}. Together, these works motivate evaluating simulation not only as an environment, but also as an instrument for preserving real-world model-selection decisions.

A related methodological issue is how to measure whether simulated evaluation preserves the conclusions that matter. Pearson correlation captures score-level linear agreement, but it can be unstable when policies have similar real-world success rates and does not directly measure whether the simulator ranks policies correctly. Rank-based metrics such as Spearman correlation better capture model-selection consistency, while Mean Maximum Rank Violation (MMRV) further penalizes rank inversions according to the real-world performance margin between mis-ranked policies~\citep{li2024simpler}. Similar concerns have been raised in large-scale robot evaluation and reinforcement-learning benchmarking, where statistically grounded comparisons and uncertainty-aware reporting are necessary to avoid over-interpreting small empirical differences~\citep{barreiros2025lbm,agarwal2021statistical}. Our use of Spearman correlation, Pearson correlation, and MMRV therefore reflects three complementary goals: preserving ordinal rankings, tracking score-level trends, and avoiding high-margin model-selection mistakes. 

Finally, our post-training protocol is related to recent work showing that pretrained generalist policies can often be adapted to new embodiments, camera viewpoints, action spaces, and task distributions with relatively small amounts of in-domain data. Octo and OpenVLA both emphasize fine-tuning as a practical mechanism for adapting broad robot-policy priors to new setups~\citep{ghosh2024octo,kim2024openvla}, while $\pi_0$ and GR00T further demonstrate that large pretrained robot policies can combine zero-shot generalization with task- or embodiment-specific adaptation~\citep{black2024pi0,bjorck2025gr00t}. Related real-world systems such as Mobile ALOHA also show that co-training with existing data can substantially improve task-specific performance from modest demonstration sets~\citep{fu2024mobilealoha}. In this context, our simulator-based and real-world co-training should be interpreted primarily as lightweight distribution calibration: the additional demonstrations adapt pretrained VLA priors to our task suite, embodiment, object placement distribution, and evaluation protocol, rather than teaching low-level manipulation skills from scratch. 

\section{Metrics}\label{metrics}

This appendix provides the detailed definitions of the metrics used to measure
sim-and-real correlation in Sec.~\ref{sec:result}. We evaluate a simulator not only by whether
its absolute success rates are close to real-world success rates, but more
importantly by whether it preserves decision-relevant conclusions, including
policy rankings and perturbation-wise failure patterns.

\subsection{Notation}

Let $\mathcal{M}=\{m_1,\ldots,m_N\}$ denote the set of evaluated policies, and
let $\mathcal{D}_e$ denote the perturbation dimensions supported by simulator
$e$. In our experiments, $\mathcal{D}_e$ may include vision, layout, language,
and behavior perturbations. For simulators that do not support a certain
perturbation dimension, that dimension is excluded when computing
perturbation-wise metrics.

For a simulator $e$, policy $m_i \in \mathcal{M}$, and perturbation dimension
$d \in \mathcal{D}_e$, we denote the simulated success rate by $S^{e}_{i,d}$.
The corresponding real-world success rate is denoted by
$S^{\mathrm{real}}_{i,d}$. Unless otherwise specified, success rates are
averaged over all evaluation tasks and rollouts under the same policy and
perturbation dimension. Larger success rates indicate better performance.

\subsection{Policy-ranking proxy metrics}

For each simulator $e$ and perturbation dimension $d$, we compare the vector of
simulated policy performances
\[
\mathbf{s}^{e}_{d}
=
\left(S^{e}_{1,d}, S^{e}_{2,d}, \ldots, S^{e}_{N,d}\right)
\]
with the corresponding real-world performance vector
\[
\mathbf{s}^{\mathrm{real}}_{d}
=
\left(S^{\mathrm{real}}_{1,d},
S^{\mathrm{real}}_{2,d},
\ldots,
S^{\mathrm{real}}_{N,d}\right).
\]
We use three complementary metrics: Spearman rank correlation, Pearson
correlation, and Mean Maximum Rank Violation.

\paragraph{Spearman rank correlation.}
Spearman rank correlation measures whether simulation preserves the relative
ordering of policies observed in the real world. It is computed as the Pearson
correlation between the ranks of the two performance vectors:
\[
\rho(e,d)
=
\mathrm{corr}
\left(
\mathrm{rank}(\mathbf{s}^{e}_{d}),
\mathrm{rank}(\mathbf{s}^{\mathrm{real}}_{d})
\right).
\]
A higher value indicates stronger ranking consistency. In particular,
$\rho=1$ means that the simulator produces exactly the same policy ordering as
the real-world evaluation, while $\rho=-1$ means that the ordering is completely
reversed. This metric is useful because policy selection is often a relative
decision: a simulator should indicate which policy is better, even if its
absolute success rates are not perfectly calibrated.

\paragraph{Pearson correlation.}
Pearson correlation measures whether simulated scores linearly track real-world
scores:
\[
r(e,d)
=
\frac{
\sum_{i=1}^{N}
\left(S^{e}_{i,d} - \bar{S}^{e}_{d}\right)
\left(S^{\mathrm{real}}_{i,d} - \bar{S}^{\mathrm{real}}_{d}\right)
}{
\sqrt{
\sum_{i=1}^{N}
\left(S^{e}_{i,d} - \bar{S}^{e}_{d}\right)^2
}
\sqrt{
\sum_{i=1}^{N}
\left(S^{\mathrm{real}}_{i,d} - \bar{S}^{\mathrm{real}}_{d}\right)^2
}
},
\]
where $\bar{S}^{e}_{d}$ and $\bar{S}^{\mathrm{real}}_{d}$ are the mean simulated
and real-world success rates across policies, respectively. A higher Pearson
correlation indicates that differences in simulated success rates are more
linearly aligned with differences in real-world success rates. Compared with
Spearman correlation, Pearson correlation is sensitive to score magnitudes and
therefore captures score-level calibration rather than only ranking consistency.

\paragraph{Mean Maximum Rank Violation.}
Spearman correlation captures ranking agreement, but it treats ranking mistakes
purely as ordinal errors and ignores the real-world performance margin between
mis-ranked policies. Reversing two policies with nearly identical real-world
success rates is less severe than reversing two policies with a large real-world
performance gap. To account for this, we use Mean Maximum Rank Violation
(MMRV), following SIMPLER~\cite{li2024simpler}.

For a fixed simulator $e$ and perturbation dimension $d$, let
\[
R_i = S^{\mathrm{real}}_{i,d},
\qquad
R_{S,i} = S^{e}_{i,d}
\]
denote the real-world and simulated success rates of policy $m_i$,
respectively. For each ordered pair of policies $(m_i,m_j)$, the pairwise rank
violation is defined as
\[
\mathrm{RankViolation}(i,j)
=
|R_i - R_j|
\cdot
\mathbf{1}
\left[
(R_{S,i} < R_{S,j}) \neq (R_i < R_j)
\right],
\]
where $\mathbf{1}[\cdot]$ is the indicator function. The indicator checks
whether the relative ordering induced by simulation disagrees with the relative
ordering observed in the real world. The violation is weighted by
$|R_i-R_j|$, so mis-ranking policies with a large real-world performance margin
is penalized more heavily than mis-ranking policies with similar real-world
performance.

The Mean Maximum Rank Violation is then computed as
\[
\mathrm{MMRV}(e,d)
=
\frac{1}{N}
\sum_{i=1}^{N}
\max_{1 \le j \le N}
\mathrm{RankViolation}(i,j).
\]
This aggregation first takes the worst rank violation associated with each
policy and then averages over policies. Lower MMRV indicates better
sim-and-real ranking consistency. In particular, $\mathrm{MMRV}=0$ means that
the simulator produces no rank violation relative to the real-world evaluation.

\paragraph{Aggregation across perturbation dimensions.}
For each simulator, we report the metric value for every perturbation dimension
and also report the average across supported dimensions:
\[
\overline{\rho}(e)
=
\frac{1}{|\mathcal{D}_e|}
\sum_{d \in \mathcal{D}_e}
\rho(e,d),
\]
with $\overline{r}(e)$ and $\overline{\mathrm{MMRV}}(e)$ computed analogously.
For correlation metrics, higher is better. For MMRV, lower is better.

\subsection{Perturbation-sensitivity metrics}

Policy ranking alone does not fully characterize simulator usefulness. A
simulator may rank policies correctly on average, but still induce unrealistic
failure modes under different perturbations. Therefore, we additionally measure
whether simulation reproduces the real-world perturbation-wise vulnerability
pattern.

For each environment $e$, policy $m_i$, and perturbation dimension $d$, we
define the normalized perturbation sensitivity as
\[
I_{e,i,d}
=
\frac{
\max_{d' \in \mathcal{D}_e} S^{e}_{i,d'}
-
S^{e}_{i,d}
}{
\max_{d' \in \mathcal{D}_e} S^{e}_{i,d'}
-
\min_{d' \in \mathcal{D}_e} S^{e}_{i,d'}
}.
\]
This normalization removes the absolute success-rate scale and focuses on the
relative severity of each perturbation for the same policy. A value close to
$0$ means that the perturbation is among the least damaging dimensions for that
policy, while a value close to $1$ means that it is among the most damaging
dimensions.

We compute the same normalized sensitivity values for the real-world
environment:
\[
I_{\mathrm{real},i,d}
=
\frac{
\max_{d' \in \mathcal{D}_e} S^{\mathrm{real}}_{i,d'}
-
S^{\mathrm{real}}_{i,d}
}{
\max_{d' \in \mathcal{D}_e} S^{\mathrm{real}}_{i,d'}
-
\min_{d' \in \mathcal{D}_e} S^{\mathrm{real}}_{i,d'}
}.
\]
Both simulated and real-world sensitivities are computed over the same set of
perturbation dimensions $\mathcal{D}_e$ so that the comparison is well defined.

\subsection{Perturbation-sensitivity alignment}

To evaluate whether simulated perturbations induce realistic vulnerability
patterns, we compare the simulated and real-world sensitivity profiles across
all policies and perturbation dimensions. We flatten the sensitivity values into
two vectors:
\[
\mathbf{i}^{e}
=
\left\{
I_{e,i,d}
\right\}_{i=1,\ldots,N;\ d \in \mathcal{D}_e},
\qquad
\mathbf{i}^{\mathrm{real}}
=
\left\{
I_{\mathrm{real},i,d}
\right\}_{i=1,\ldots,N;\ d \in \mathcal{D}_e}.
\]
We then compute three alignment metrics.

\paragraph{Sensitivity Spearman correlation.}
Sensitivity Spearman correlation measures whether simulation preserves the
relative ordering of perturbation severity:
\[
\rho_{\mathrm{sens}}(e)
=
\mathrm{corr}
\left(
\mathrm{rank}(\mathbf{i}^{e}),
\mathrm{rank}(\mathbf{i}^{\mathrm{real}})
\right).
\]
A higher value means that perturbations ranked as more damaging in the real
world are also ranked as more damaging in simulation.

\paragraph{Sensitivity Pearson correlation.}
Sensitivity Pearson correlation measures whether the normalized sensitivity
values are linearly aligned:
\[
r_{\mathrm{sens}}(e)
=
\mathrm{corr}
\left(
\mathbf{i}^{e},
\mathbf{i}^{\mathrm{real}}
\right).
\]
A higher value indicates that the simulator not only preserves the ordering of
perturbation severity, but also matches the relative magnitude of vulnerability
across perturbations.

\paragraph{Sensitivity MAE.}
Finally, we report the mean absolute error between simulated and real-world
sensitivity values:
\[
\mathrm{MAE}_{\mathrm{sens}}(e)
=
\frac{1}{N|\mathcal{D}_e|}
\sum_{i=1}^{N}
\sum_{d \in \mathcal{D}_e}
\left|
I_{e,i,d}
-
I_{\mathrm{real},i,d}
\right|.
\]
Lower sensitivity MAE indicates that the simulator more accurately reproduces
the real-world severity of each perturbation. Together, sensitivity Spearman,
sensitivity Pearson, and sensitivity MAE measure whether a simulator makes
policies fail in ways that are consistent with real-world deployment.

Overall, the policy-ranking metrics evaluate whether simulation can support
model selection, while the perturbation-sensitivity metrics evaluate whether
simulation can support failure diagnosis. Spearman correlation focuses on
relative ordering, Pearson correlation focuses on score-level linear alignment,
MMRV focuses on the severity of high-margin ranking mistakes, and sensitivity
MAE measures the absolute discrepancy between simulated and real-world
vulnerability profiles. Therefore, a reliable simulator should achieve high
Spearman and Pearson correlations, low MMRV, high sensitivity correlations, and
low sensitivity MAE.


\section{Details of post-training recipe}

\subsection{Data collection}

We collect all task demonstrations using a Meta Oculus Quest 2-based teleoperation pipeline. The headset and controllers provide an intuitive interface for recording human demonstrations, allowing operators to control the robot end-effector and gripper while observing the task scene in real time. To facilitate reproducibility and future research, we will release all data collected in this process together with the corresponding teleoperation code. Notably, the original DROID embodiment configuration used by GR00T adopts an eef-9D action/state modality, which is not directly aligned with the data produced by our teleoperation pipeline. Therefore, when training or fine-tuning GR00T models on our collected demonstrations, we reuse the new DROID embodiment defined in Sec.~\ref{configs}, which represents the robot state and action using joint positions and gripper states. This ensures that the GR00T input-output modalities are consistent with our data format.

\subsection{Training configs}
\label{configs}

To reproduce the results, we provide training configs for all baselines on our dataset. The same hyperparameters are used as below, and learning rates are set as default:

\begin{table}[h]
\begin{tabular}{c|ccccc}
\toprule
Models      & Batchsize & Training steps & Action horizon & base model & Embodiment \\
\midrule
$\pi$       & 32        & 10,000         & 16              & DROID      & -
\\       
GR00T       & 32        & 10,000         & 16              & DROID      & New         \\
\bottomrule
\end{tabular}
\end{table}

Example for fine-tuning $\pi_0$-fast:

\begin{lstlisting}[language=Python]

TrainConfig(
        name="pi0_fast_droid_finetune",
        model=pi0_fast.Pi0FASTConfig(
            action_dim=8,
            action_horizon=16,
            max_token_len=180,
            paligemma_variant="gemma_2b_lora",
        ),
        data=LeRobotDROIDDataConfig(
            base_config=DataConfig(prompt_from_task=True),
            assets=AssetsConfig(
                assets_dir="",
                asset_id="",
            ),
        ),
        weight_loader=weight_loaders.CheckpointWeightLoader(""),
        num_train_steps=10_000,
        batch_size=32,
        freeze_filter=pi0_fast.Pi0FASTConfig(
            action_dim=8,
            action_horizon=16,
            max_token_len=180,
            paligemma_variant="gemma_2b_lora",
        ).get_freeze_filter(),
        ema_decay=None, 
    ),


\end{lstlisting}

Below we give the registration of DROID embodiment (align with our data) for GR00T models.

\begin{lstlisting}[language=Python]
VIDEO_KEYS = ["image", "wrist_image"]
STATE_KEYS = ["joint_position", "gripper"]
ACTION_KEYS = ["joint_position", "gripper"]
LANGUAGE_KEYS = ["annotation.human.action.task_description"]
ACTION_HORIZON = 16


new_droid_config = {
    "video": ModalityConfig(
        delta_indices=[0],
        modality_keys=VIDEO_KEYS,
    ),
    "state": ModalityConfig(
        delta_indices=[0],
        modality_keys=STATE_KEYS,
    ),
    "action": ModalityConfig(
        delta_indices=list(range(ACTION_HORIZON)),
        modality_keys=ACTION_KEYS,
        action_configs=[
            ActionConfig(
                rep=ActionRepresentation.ABSOLUTE,
                type=ActionType.NON_EEF,
                format=ActionFormat.DEFAULT,
                state_key="joint_position",
            ),
            ActionConfig(
                rep=ActionRepresentation.ABSOLUTE,
                type=ActionType.NON_EEF,
                format=ActionFormat.DEFAULT,
                state_key="gripper",
            ),
        ],
    ),
    "language": ModalityConfig(
        delta_indices=[0],
        modality_keys=LANGUAGE_KEYS,
    ),
}


register_modality_config(new_droid_config, embodiment_tag=EmbodimentTag.NEW_EMBODIMENT)
\end{lstlisting}

\section{Perturbation settings}\label{pertvis}

For some simulators, not all perturbation dimensions are natively provided by the original benchmark. To enable a fair cross-simulator comparison, we supplement the missing perturbations using comparable perturbation protocols whenever possible. Specifically, for vision, layout, and language perturbations, we apply perturbations with similar semantic meaning and difficulty levels across simulators, so that the resulting evaluation reflects differences in simulator behavior rather than differences in perturbation availability. The only exception is the unseen-object dimension. Since object assets, geometry, affordances, and task-specific object sets are difficult to make identical across different simulators, enforcing a shared unseen-object protocol would introduce additional confounding factors. Therefore, for this dimension, we follow each simulator's default unseen-object setting when available. Since SIMPLER does not include an unseen-object perturbation setting, we exclude this dimension when reporting its perturbation-wise results.

For language perturbations, we use shared textual perturbation templates across simulators to ensure comparable instruction-level shifts. Although the textual changes are controlled to be similar, their effects are not purely linguistic. For VLA policies, an instruction must be grounded in simulator-specific visual observations, task instantiations, object appearances, action dynamics, and success conditions. Therefore, language perturbation does not measure language understanding in isolation, but rather the robustness of the full vision-language-action grounding loop under changed instructions. From this perspective, applying language perturbations with comparable intensity across simulators is reasonable: it allows us to test whether each simulator induces consistent instruction-conditioned policy behavior under similar language shifts.

It is worth noting that the original layout perturbation protocol in REALM is not fully aligned with the layout perturbations used in the other simulators. In the original REALM setting, the positions of the manipulated objects remain fixed, while the layout variation is mainly introduced by changing the appearance or placement of non-manipulated distractor objects. We consider this perturbation to be closer to background noise than to a true layout shift. If the manipulated object always appears at the same location, it is difficult to determine whether the policy has learned the task semantics and spatial grounding, or whether it simply reproduces a memorized trajectory from the default scene configuration. Therefore, we improve the original layout perturbation by varying the placement of the manipulated objects, so that the evaluation more directly tests the policy's spatial generalization ability under changed task layouts.

\begin{figure}
    \centering
    \includegraphics[width=0.7\linewidth]{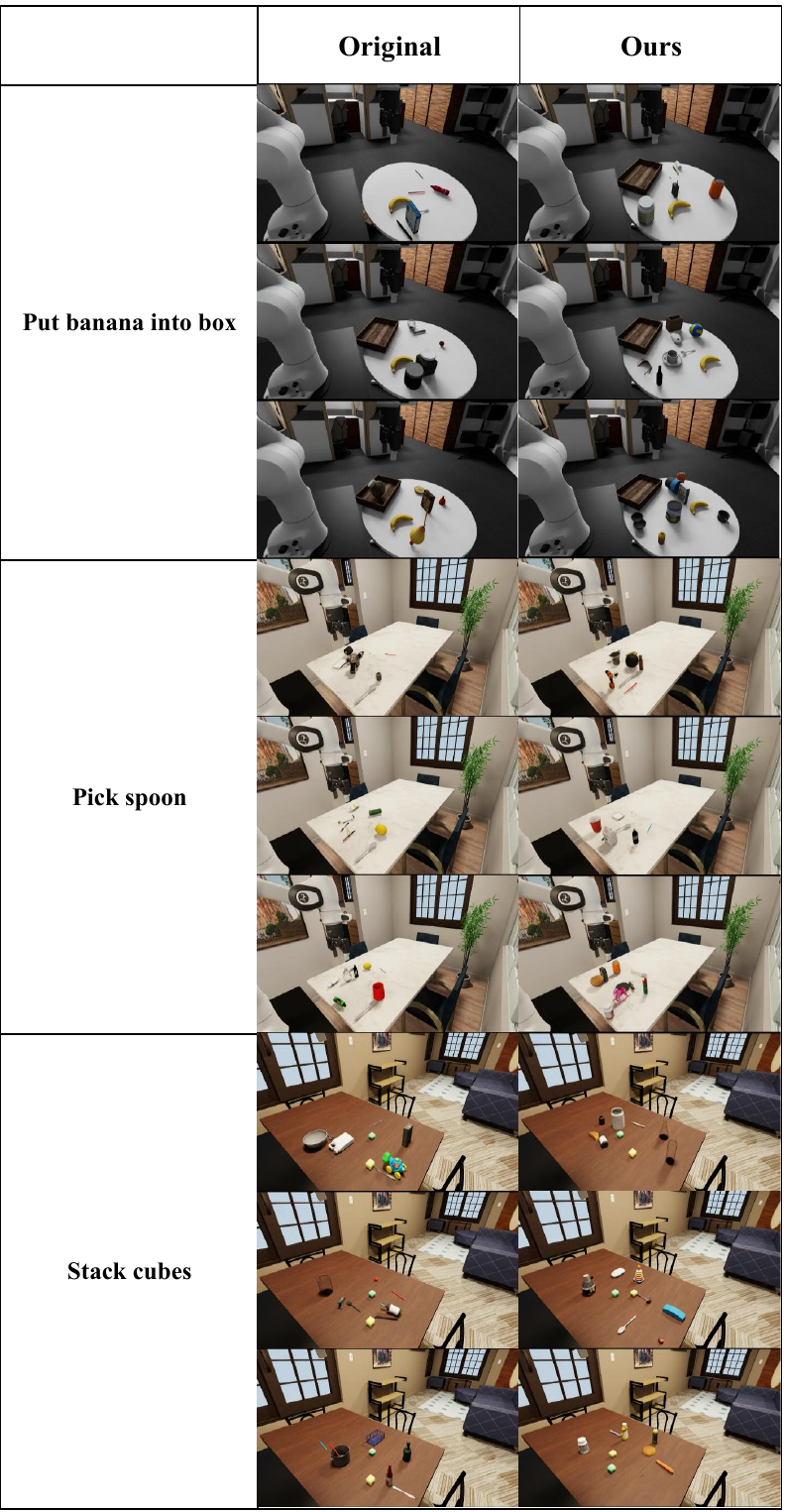}
    \caption{The comparison of original and new layout perturbation}
    \label{layout}
\end{figure}

Correspondingly, during data collection, we also adjust the object placement distribution to expose the policy to moderate layout variations in the target task setting. Instead of collecting demonstrations only from a fixed manipulated-object location, we sample different feasible object placements while preserving the same task objective and success criterion. This allows the model to learn the layout-dependent action mapping during post-training, rather than overfitting to a single canonical trajectory. The difference between the original and improved layout perturbation protocols is visualized in Fig.~\ref{layout}, where the improved setting explicitly changes the manipulated-object positions and therefore provides a more faithful test of layout-level generalization.

\end{document}